\documentclass[11pt,letterpaper]{article}
\usepackage{naaclhlt2015}
\usepackage{times}
\usepackage{latexsym}
\usepackage{graphicx}
\usepackage{algorithm2e}
\usepackage{booktabs}
\usepackage{paralist}
\usepackage{color}
\usepackage{pdfsync}
\newcommand{\ra}[1]{\renewcommand{\arraystretch}{#1}}

\usepackage{xspace,multirow}

\newcommand{\z}{\phantom{0}}
\newcommand{\np}{---}
\newcommand{\nr}{???}

\newcommand{\kd}{$k$-d\xspace}

\newcommand{\tabref}[2][]{Table#1~\ref{#2}\xspace}

\newcommand{\dataset}[1]{\textsc{#1}\xspace}
\newcommand{\GEOTEXT}{\dataset{GeoText}}
\newcommand{\TwitterUS}{\dataset{Twitter-US}}
\newcommand{\TwitterWorld}{\dataset{Twitter-World}}

\newcommand{\method}[1]{\texttt{#1}\xspace}
\newcommand{\LR}{\method{LR}}
\newcommand{\LP}{\method{LP}}
\newcommand{\LPLR}{\method{LP-LR}}

\newcommand{\nearacc}{Acc@161\xspace}
\newcommand{\Mean}{Mean\xspace}
\newcommand{\Median}{Median\xspace}

\newcommand{\cisaff}{\ensuremath{1}\xspace}
\newcommand{\mathaff}{\ensuremath{2}\xspace}

\title{Exploiting Text and Network Context for Geolocation of Social
  Media Users}

\author{Afshin Rahimi,$^{\cisaff}$ Duy Vu,$^{\mathaff}$ Trevor Cohn,$^{\cisaff}$ \and Timothy Baldwin$^{\cisaff}$\\
  $^{\cisaff}$Department of Computing and Information Systems\\
  $^{\mathaff}$Department of Mathematics and Statistics\\
        The University of Melbourne\\
  \texttt{arahimi@student.unimelb.edu.au}\\
  \texttt{\{duy.vu, t.cohn\}@unimelb.edu.au}\\
  \texttt{tb@ldwin.net}}

\date{}

\begin{document}

\maketitle

\begin{abstract}
  Research on automatically geolocating social media users 
  has conventionally been based on the text content of posts from a
  given user or the social network of the user, with very little
  crossover between the two, and no benchmarking of the two approaches
  over comparable datasets. We bring the two threads of research
  together in first proposing a text-based method based on
  adaptive grids,
  followed by a hybrid network- and text-based method. 
  Evaluating over
  three Twitter datasets, we show that the empirical difference between
  text- and network-based methods is not great, and that hybridisation
  of the two is superior to the component methods, especially in contexts
  where the user graph is not well connected. We achieve
  state-of-the-art results on all three datasets.
\end{abstract}

\section{Introduction}

There has recently been a spike in interest in the task of inferring the
location of users of social media services, due to its utility in
applications including location-aware information retrieval~\cite{amitay2004web}, recommender
systems~\cite{noulas2012random} and rapid disaster response~\cite{earle2010omg}.
Social media sites such as Twitter and Facebook provide two primary
means for users to declare their location: (1) through text-based
metadata fields in the user's profile; and (2) through GPS-based
geotagging of posts and check-ins.  However, the text-based metadata is
often missing, misleading or imprecise, and only a tiny proportion of users
geotag their posts~\cite{cheng2010you}.
Given the small number of users with reliable location information,
there has been significant interest in the task of automatically
geolocating (predicting lat/long coordinates) of users based on their publicly
available posts, metadata and social network information. These
approaches are built on the premise that a user's location is evident
from their posts, or through location homophily in their social network. 

Our contributions in this paper are: 
\begin{inparaenum}[\itshape a\upshape)]
\item the demonstration that
network-based methods are generally superior to text-based user
geolocation methods due to their robustness; 
\item the proposal of a hybrid classification method
that backs-off from network- to text-based predictions for disconnected
users, which we show to achieve state-of-the-art accuracy over all
Twitter datasets we experiment with; and 
\item empirical evidence to suggest that text-based geolocation methods
  are largely competitive with network-based methods.
\end{inparaenum}

\section{Related Work}
\label{sec:related}

Past work on user geolocation falls broadly into two categories:
text-based and network-based methods.
Common to both methods is the manner of framing the
geolocation prediction problem. Geographic coordinates 
are real-valued, and accordingly this is most naturally modelled as
(multiple) regression. However for modelling convenience,
the problem is typically simplified to classification by first
pre-partitioning the regions  into discrete sub-regions using 
either known city locations \cite{bo2012geolocation,rout2013s} or a
\kd tree partitioning \cite{roller2012supervised,wing2014hierarchical}. 
In the \kd tree methods, the resulting discrete regions are treated
either as a flat list (as we do here) or a nested hierarchy.

\subsection{Text-based Geolocation}

Text-based approaches assume that language in social
media is geographically biased, which is clearly evident for 
regions speaking different languages \cite{han2014text}, but is also reflected in regional dialects and the use of region specific terminology.
Text based models have predominantly used bag of words features to learn 
per-region classifiers~\cite{roller2012supervised,wing2014hierarchical},
including feature selection for location-indicative terms~\cite{bo2012geolocation}.


Topic models have also been applied to model geospatial text
usage~\cite{eisenstein2010latent,ahmed2013hierarchical}, by associating
latent topics with locations. This has a benefit of allowing for
prediction over continuous space, i.e., without the need
to render the problem as classification. On the other hand, these
methods have high algorithmic complexity and their generative
formulation is unlikely to rival the performance of discriminative
methods on large datasets.

\subsection{Network-based Geolocation} 

Although online social networking sites allow for global interaction,
users tend to befriend and interact with many of the same people online
as they do off-line~\cite{rout2013s}.  Network-based methods exploit
this property to infer the location of users from the locations of their
friends~\cite{jurgens2013s,rout2013s}.  This relies on some form of
friendship graph, through which location information can be propagated,
e.g., using label propagation \cite{jurgens2013s,talukdar2009}. A
significant caveat regarding the generality of these techniques is that
friendship graphs are often not accessible, e.g., secured from the
public (Facebook) or hidden behind
a heavily rate-limited API (Twitter).\\

While the raw accuracies reported for network-based methods (e.g.,
\newcite{jurgens2013s} and \newcite{rout2013s}) are generally higher
than those reported for text-based methods (e.g.,
\newcite{wing2014hierarchical} and \newcite{han2014text}), they
have been evaluated over different datasets and spatial representations,
making direct comparison uninformative. Part of our contribution in this
paper is direct comparison between the respective methods over standard
datasets. In this, we propose both text- and network-based methods,
and show that they achieve state-of-the-art results on three 
pre-existing Twitter geolocation corpora. We also propose a new hybrid 
method incorporating both textual and network information, which 
also improves over the state-of-the-art, and outperforms the text-only 
or network-only methods over two of the three datasets.

\section{Data}
\label{sec:data}

We evaluate on three Twitter corpora, each of which uses geotagged tweets to derive
a geolocation for each user. Each user is represented by the concatenation of
their tweets, and is assumed to come from a single location. 

\begin{description}
\item{\GEOTEXT:} around 380K tweets from 9.5K users based in
contiguous USA, of which 1895 is held out for development and testing
~\cite{eisenstein2010latent}; the location of
each user is set to the GPS coordinates of their first tweet.
\item{\TwitterUS:} around 39M tweets from 450K users based in the
contiguous USA. 10K users are held out for each of development and
testing~\cite{roller2012supervised}; again users' locations are taken from their
first tweet.
\item{\TwitterWorld:} around 12M English tweets from 1.4M users
based around the world, of which 10K users are held out for each of development and testing
~\cite{bo2012geolocation}; users are geotagged with the
centre of the closest city to their tweets.
\end{description}

In each case, we use the established training, development and testing
partitions, and follow \newcite{cheng2010you} and
\newcite{eisenstein2010latent} in evaluating based on: (1) accuracy at
161km~(``\nearacc''); (2) mean error distance, in kilometres
(``\Mean''); and (3) median error distance, in kilometres (``\Median'').

\section{Methods}

\subsection{Text-based Classification} 

Our baseline method for text based geolocation is based
on~\newcite{wing2014hierarchical}, who formulate the geolocation problem
as classification using \kd trees. In summary, their approach first
discretises the continuous space of geographical coordinates using a \kd
tree such that each sub-region (leaf) has similar numbers of users. This
results in many small regions for areas of high population density and
fewer larger regions for country areas with low population density.
Next, they use these regions as class labels to train a logistic
regression model (``\LR''). Our work is also subject to a sparse $l_1$ regularisation
penalty~\cite{tibshirani1996regression}. In their work,
\newcite{wing2014hierarchical} showed that hierarchical logistic
regression with a beam search achieves higher results than logistic
regression over a flat label set, but in this research, we use a 
flat representation, and leave experiments with hierarchical
classification to future work.


For our experiments, the number of users in each region was selected
from $\{300, 600, 900, 1200\}$ to optimise median error distance on
the development set, resulting in values of 300, 600 and 600 for
\GEOTEXT, \TwitterUS and \TwitterWorld, respectively. The $l_1$
regularisation coefficient was also optimised  in the same manner.

As features, we used a bag of unigrams (over both words and @-mentions) and removed all
features that occurred in fewer than 10 documents,
following~\newcite{wing2014hierarchical}. The features for each user were
weighted using tf-idf, followed by per-user $l_2$ normalisation.
The normalisation is particularly important because our `documents' are formed from all the tweets of each user, which
vary significantly in size between users; furthermore, this adjusts for
differing degrees of lexical variation~\cite{lee1995combining}. 
The number of features was almost 10K for \GEOTEXT and about
2.5M for the other two corpora. 
For evaluation we use the median of all training locations in the
sub-region predicted by the classifier, from which we measure the error
against a test user's gold standard location.

\subsection{Network-based Label Propagation}

Next, we consider the approach of~\newcite{jurgens2013s} who
used label propagation (``\LP''; \newcite{zhu2002learning}) to infer user locations
using social network structure. \newcite{jurgens2013s} defined an
undirected network from interactions among Twitter users based on @-mentions in their tweets,
a mechanism typically used for conversations between
friends. Consequently these links often correspond to offline
friendships, and accordingly the network will exhibit a high degree of
location homophily. The network is constructed by defining as nodes
all users in a dataset (train and test), as well as other
external users mentioned in their tweets. Unlike~\newcite{jurgens2013s}
who only created edges when both users mentioned one another, we created edges
if either user mentioned the other. For the three datasets used in our experiments,
bi-directional mentions were too rare to be useful, and we thus used the (weaker) uni-directional 
mentions as undirected edges instead.
The edges between users are 
undirected and weighted by the number of @-mentions in tweets by 
either user.\footnote{As our datasets don't have tweets for external
  users, these nodes do not contribute to the weight of their incident edges.}

The mention network statistics for each of our datasets is shown in
\tabref{tab:datasets}.\footnote{Note that @-mentions were removed in
  the published \TwitterUS and \TwitterWorld datasets. To recover
  these we rebuilt the corpora from the Twitter archive.}
Following~\newcite{jurgens2013s}, we ran the label propagation algorithm to
update the location of each non-training node to the 
weighted median of its neighbours. This process continues iteratively
until convergence, which occurred at or before 10 iterations.



\subsection{A Hybrid Method} 

Unfortunately many test users are not transitively connected
to any training node (see \tabref{tab:datasets}), meaning that \LP fails
to assign them any location. This can happen when users
don't use @-mentions, or when a set of nodes constitutes a
disconnected component of the graph.

\begin{table}
\begin{center}
\resizebox{\linewidth}{!}{%
\begin{tabular}{cccc}
\toprule
 & \GEOTEXT& \TwitterUS &\TwitterWorld\\
\midrule
User mentions &109K&3.63M&16.8M \\
Disconnected & \multirow{2}{*}{23.5\%}&\multirow{2}{*}{27.7\%}& \multirow{2}{*}{2.36\%}\\
test users:\\
\bottomrule
\end{tabular}
}
\end{center}
\caption{The graph size and proportion of test users disconnected 
  from training users for each dataset. }
\label{tab:datasets}
\end{table}

\begin{table*}\centering
\ra{1.1}
\smallskip\noindent
\resizebox{\linewidth}{!}{%
\begin{tabular}{@{}lccccccccccc@{}}\toprule
& \multicolumn{3}{c}{\GEOTEXT} & \phantom{abc}& \multicolumn{3}{c}{\TwitterUS} &
\phantom{abc} & \multicolumn{3}{c}{\TwitterWorld}\\
\cmidrule{2-4} \cmidrule{6-8} \cmidrule{10-12}
                          & \nearacc & \Mean      & \Median    &&  \nearacc & \Mean    & \Median    &&  \nearacc & \Mean      & \Median    \\ 
\midrule
\LR (text-based)           & 38.4     & 880.6      & 397.0      &&  50.1     & 686.7    & 159.2      && \bf{63.8} &\bf{\z866.5}& \bf{\z19.9} \\
\LP (network-based)        & 45.1     & 676.2      & 255.7      &&  37.4     & 747.8    & 431.5      && 56.2      & 1026.5     & \z79.8      \\
\LPLR (hybrid)             & \bf{50.2}& \bf{653.9} & \bf{151.2} &&  \bf{50.2}&\bf{620.0}& \bf{157.1} && 59.2      & \z903.6    & \z53.7      \\
\midrule
\newcite{wing2014hierarchical} (uniform)& \np   & \np        & \np        &&  49.2     & 703.6    & 170.5      && 32.7      & 1714.6     & 490.0       \\
\newcite{wing2014hierarchical} (\kd) & \np      & \np        & \np        &&  48.0     & 686.6    & 191.4      &&  31.3     & 1669.6     & 509.1       \\
\newcite{bo2012geolocation}        & \np      & \np        & \np        && 45.0      & 814      & 260        && 24.1      & 1953       & 646         \\ 
\newcite{ahmed2013hierarchical}      & \nr      & \nr        & 298        && \np       & \np      & \np        && \np       & \np        & \np         \\
\bottomrule
\end{tabular}
}
\caption{Geolocation accuracy over the 
  three Twitter corpora comparing Logistic Regression (\LR), Label Propagation (\LP)
  and LP over LR initialisation (\LPLR)
  with the state-of-the-art methods for the respective datasets (``\np''
  signifies that no results were published for the given dataset, and
  ``\nr'' signifies that no results were reported for the given metric).
}
\label{tab:results}
\end{table*}

In order to alleviate this problem,
we use the text for each test user in order to estimate their
location, which is then used as an initial estimation during label propagation.
In this hybrid approach, we first estimate the location for each test
node using the \LR classifier described above, before running label propagation
over the mention graph. This iteratively adjusts the locations based
on both the known training users and guessed test users, while simultaneously
inferring locations for the external users. In such a way, the
inferred locations of test users will better match neighbouring
users in their sub-graph, or in the case of disconnected nodes, will
retain their initial classification estimate.


\section{Results}

\tabref{tab:results} shows the performance of the three methods over the
test set for the three datasets.
The results are also compared with the state of the art for \TwitterUS and
\TwitterWorld~\cite{wing2014hierarchical}, and \GEOTEXT~\cite{ahmed2013hierarchical}.

Our methods achieve a sizeable improvement over the previous state of
the art for all three datasets. \LPLR performs best over \GEOTEXT and
\TwitterUS, while \LR performs best over \TwitterWorld; the reduction in
median error distance over the state of the art ranges from around 40\%
to over 95\%. Even for \TwitterWorld, the results for \LPLR are
substantially better than the best-published results for that dataset.

Comparing \LR and \LP, no strong conclusion can be drawn --- the
text-based \LP actually outperforms the network-based \LR for two of the
three datasets, but equally, the combination of the two (\LPLR) performs
better than either component method over two of the three datasets. For
the third (\TwitterWorld), \LR outperforms \LPLR due to a combination of
factors. First, unlike the other two datasets, the label set is
pre-discretised (everything is aggregated at the city level), meaning
that \LP and \LR use the same label set.\footnote{For consistency, we
  learn a \kd tree for \TwitterWorld and use the merged representation
  for \LR, but the \kd tree largely preserves the pre-existing city
  boundaries.}  This annuls the representational advantage that \LP has
in the case of the other two datasets, in being able to capture a more
fine-grained label set (i.e., all locations associated with training
users). Second, there are substantially fewer disconnected test users in
\TwitterWorld (see \tabref{tab:datasets}), meaning that the results for
the hybrid \LPLR method are dominated by the empirically-inferior \LP. 

Although \LR is similar to \newcite{wing2014hierarchical}, we achieved
large improvements over their reported results. This might be due to:
(a) our use of @-mention features; (b) $l_1$ regularisation, which is essential to
preventing overfitting for large feature sets; or (c) our use of $l_2$
normalisation of rows in the design matrix, which we found reduced
errors by about 20\% on \GEOTEXT, in keeping with results from text
categorisation~\cite{lee1995combining}. Preliminary experiments also
showed that 
lowering the term frequency threshold from 10 can further improve the
\LR results on all three datasets.

\LP requires few hyper-parameters and is relatively robust.  It
converged on all datasets in fewer than 10 iterations, and geolocates not
only the test users but all nodes in the mention graph.  Another
advantage of \LP over \LR is the relatively modest amount of memory and
processing power it requires.

\section{Conclusion}

We proposed a series of approaches to social media user geolocation
based on: (1) text-based analysis using logistic regression with
regularisation; (2) network-based analysis using label propagation; and
(3) a hybrid method based on network-based label propagation, and
back-off to text-based analysis for disconnected users. We achieve
state-of-the-art results over three pre-existing Twitter datasets, and
find that, overall, the hybrid method is superior to the two component
methods.The \LPLR method is a hybrid approach that uses the \LR predictions as priors.
It is not simply a backoff from network information to textual information in the sense that
it propagates the \LR geolocations through the network.
That is, if a test node is disconnected from the training nodes but still has connections to other test nodes,
the geolocation of the node is adjusted and propagated through the network. 
It is possible to add extra nodes to the graph after applying the
algorithm and to geolocate only these nodes efficiently, although this
approach is potentially less accurate than inferencing over the full graph from scratch. 

Label propagation algorithms such as Modified Adsorption~\cite{talukdar2009} allow for different levels of influence between 
prior/known labels and propagated label distributions. These algorithms require a discretised output space for label propagation, 
while \LP can work directly on continuous data. We leave label propagation over discritised output and allowing different influence
levels between prior and propagated label distributions to future work.

There is no clear consensus on whether text- or
network-based methods are empirically superior at the user geolocation task.
Our results show that the network-based method (\LP) is more robust than the text-based (\LR) method as it 
requires a smaller number of hyper-parameters, uses less memory and computing resources, converges much faster and 
geolocates not only test users but all mentioned users. The drawback of \LP is that it fails to geolocate disconnected
test users. So for connected nodes -- the majority of test nodes in all our datasets -- \LP is more robust than \LR.
Text-based methods are very sensitive to the regularisation settings and the types of textual features. That said, with 
thorough parameter tuning, they might outperform network-based method in terms of accuracy.

In future work, we hope to look at different types of network
information for label propagation, more precise propagation methods to
deal with non-local interactions, and also efficient ways of utilising
both textual and network information in a joint model.



\subsection*{Acknowledgements}

We thank the anonymous reviewers for their insightful comments and
valuable suggestions. This work was funded in part by the Australian
Research Council.

\bibliographystyle{naaclhlt2015}
\bibliography{Master}
\end{document}